\documentclass[10pt,conference]{IEEEtran}
\IEEEoverridecommandlockouts
\usepackage{cite}
\usepackage{amsmath,amssymb,amsfonts}
\usepackage{algorithmic}
\usepackage{graphicx}
\usepackage{textcomp}
\newcommand{\bW}{\boldsymbol{W}}
\newcommand{\bx}{\boldsymbol{x}}
\newcommand{\by}{\boldsymbol{y}}
\newcommand{\bu}{\boldsymbol{u}}
\usepackage{xcolor}
\usepackage{subcaption}
\usepackage[utf8]{inputenc} 
\usepackage[T1]{fontenc}    

\usepackage[english]{babel} 

\usepackage{setspace} 
\usepackage{amsmath}
\usepackage{amssymb}
\usepackage{amsthm} 
\usepackage{bm}
\usepackage{listofitems} 

\usepackage{graphicx}
\graphicspath{{figures/}} 
\usepackage{caption}      
\usepackage{subcaption}   
\usepackage{tikz}
\usetikzlibrary{arrows.meta, matrix} 

\usepackage{booktabs} 
\usepackage{longtable} 
\usepackage{tabularx} 
\usepackage{multirow}
\usepackage{multicol}
\usepackage{diagbox}  

\usepackage{xcolor}
\definecolor{darkgreen}{rgb}{0.0, 0.5, 0.0}
\colorlet{myred}{red!80!black}
\colorlet{myblue}{blue!80!black}
\colorlet{mygreen}{green!60!black}
\colorlet{myorange}{orange!70!red!60!black}
\colorlet{mydarkred}{red!30!black}
\colorlet{mydarkblue}{blue!40!black}
\colorlet{mydarkgreen}{green!30!black}

\usepackage[a-2b,latxmp]{pdfx}

\hypersetup{
    colorlinks=true,
    linkcolor=black,
    citecolor=black,
    urlcolor=blue,
    pdfborder={0 0 0},
    breaklinks=true,
}

\usepackage{listings}

\lstdefinestyle{python_style}{
    language=Python,
    basicstyle=\ttfamily\tiny,
    keywordstyle=\color{blue},
    stringstyle=\color{red},
    commentstyle=\color{green!50!black},
    numberstyle=\tiny\color{gray},
    frame=single, 
    breaklines=true, 
    tabsize=4,
    captionpos=b, 
    xleftmargin=5pt, 
    xrightmargin=5pt
}

\usepackage[ruled, vlined, linesnumbered]{algorithm2e}

\usepackage{fancyhdr}
\fancypagestyle{plain}{ 
  \fancyhf{} 
  \fancyfoot[C]{\thepage} 

}

\usepackage[capitalize]{cleveref} 

\tikzset{
  >=latex, 
  node/.style={thick,circle,draw=myblue,minimum size=22,inner sep=0.5,outer sep=0.6},
  node in/.style={node,green!20!black,draw=mygreen!30!black,fill=mygreen!25},
  node hidden/.style={node,blue!20!black,draw=myblue!30!black,fill=myblue!20},
  node convol/.style={node,orange!20!black,draw=myorange!30!black,fill=myorange!20},
  node out/.style={node,red!20!black,draw=myred!30!black,fill=myred!20},
  connect/.style={thick,mydarkblue}, 
  connect arrow/.style={-{Latex[length=4,width=3.5]},thick,mydarkblue,shorten <=0.5,shorten >=1},
  node 1/.style={node in}, 
  node 2/.style={node hidden},
  node 3/.style={node out}
}

\usepackage{threeparttable}

\def\BibTeX{{\rm B\kern-.05em{\sc i\kern-.025em b}\kern-.08em
    T\kern-.1667em\lower.7ex\hbox{E}\kern-.125emX}}

\usepackage{censor}
\StopCensoring

\begin{document}

\title{Quantization-Aware Regularizers for Deep Neural Networks Compression 
\thanks{\blackout{M.F.\ and D.M.\  have been partially supported by the Italian MUR PRIN project ``Multicriteria data structures and algorithms: from compressed to learned indexes, and beyond'' (Prot. 2017WR7SHH). Additional support to D.M. has been granted by the National Plan for NRRP Complementary Investments (PNC) in the call for the funding of research initiatives for technologies and innovative trajectories in the health—project n. PNC0000003—AdvaNced Technologies for Human-centrEd Medicine (project acronym: ANTHEM).
}
}
}

\author{\IEEEauthorblockN{1\textsuperscript{st} \censor{Dario Malchiodi}}
\IEEEauthorblockA{\textit{\censor{Dipartimento di Informatica} 
} \\
\textit{\censor{Università degli Studi di Milano}}\\
\censor{Milan, Italy} \\
\censor{dario.malchiodi@unimi.it}}
\and
\IEEEauthorblockN{2\textsuperscript{nd} \censor{Mattia Ferraretto}}
\IEEEauthorblockA{\textit{\censor{Dipartimento di Informatica} 
} \\
\textit{\censor{Università degli Studi di Milano}}\\
\censor{Milan, Italy} \\
\censor{ferraretto.mattia@proton.me}}
\and
\IEEEauthorblockN{3\textsuperscript{rd} \censor{Marco Frasca}}
\IEEEauthorblockA{\textit{\censor{Dipartimento di Informatica} 
} \\
\textit{\censor{Università degli Studi di Milano}}\\
\censor{Milan, Italy} \\
\censor{marco.frasca@unimi.it}}
}

\maketitle

\begin{abstract}
Deep Neural Networks reached state‑of‑the‑art performance across numerous domains, but this progress has come at the cost of increasingly large and over‑parameterized models, posing serious challenges for deployment on resource‑constrained devices. As a result, model compression has become essential, and---among compression techniques---weight quantization is largely used and particularly effective, yet it typically introduces a non‑negligible accuracy drop. However, it is usually applied to already trained models, without influencing how the parameter space is explored during the learning phase. 
In contrast, we introduce per‑layer regularization terms that drive weights to naturally form clusters during training, integrating quantization awareness directly into the optimization process. This reduces the accuracy loss typically associated with quantization methods  while preserving their compression potential. Furthermore, in our framework quantization representatives become network parameters, marking, to the best of our knowledge, the first approach to embed quantization parameters directly into the backpropagation procedure. Experiments on CIFAR‑10 with AlexNet and VGG16 models confirm the effectiveness of the proposed strategy.
\end{abstract}


\section{Introduction}
Over the past decade, Deep Neural Networks (DNNs) have achieved state‑of‑the‑art performance across a wide range of complex tasks, spanning from computer vision~\cite{He16} to natural language processing~\cite{Vaswani17}.
This rapid progress has come at the cost of exponentially increasing model complexity, with modern architectures often containing millions or even billions of parameters.
Yet, despite this growth in scale, trained networks commonly exhibit substantial parameter redundancy~\cite{Denil13}.
Such over-parameterization leads to large memory footprints, high energy consumption, and increased computational latency, which makes deploying powerful models on resource-limited edge devices increasingly infeasible. As a result, DNN compression has become essential, aiming to lower the resource requirements of learned models~\cite{Guo16,deep_survey}. 
Prominent techniques include connection pruning~\cite{Guo16}, filter pruning~\cite{Li17}, weight and activation quantization~\cite{Hubara17}, low-rank factorization~\cite{Denton14}, and knowledge distillation~\cite{Hinton15}.
In particular, \emph{weight quantization} denotes reducing the number of bits used in order to represent network weights, for instance by lowering floating-point precision~\cite{Tung20}, down to the extreme of a single bit~\cite{Courbariaux15}.
Weights can be quantized either by directly mapping them to the nearest value in a deterministic or stochastic manner~\cite{Hubara17}, or by optimizing a given metric, such as entropy~\cite{Park17}. Resource savings can also be achieved \textit{indirectly} through weight sharing, 
where only a small set of full‑precision representative weights is stored and reused across multiple weight clusters~\cite{Han15,deep_survey,marino_comp_strat, marino_sham}. In this case, compression is obtained by employing suitable lossless encoding schemes that exploit the low‑entropy distribution induced by weight quantization~\cite{Han15,efficient_entropy_coding}. 
Most existing methods, however, perform such quantization after standard gradient‑based updates, even when quantization is used during training~\cite{Tung20}. 
Only a limited number
 of works incorporate a quantization‑aware objective directly into the loss, and these typically alternate between steps dedicated to accuracy and steps dedicated to compression~\cite{Zhang20}.

In this work, we propose per-layer weight regularization loss terms specifically designed to drive weights to `naturally' form clusters during DNN training, directly via gradient computation and without any dedicated post-update steps.  
Notably, both previously discussed weight quantization families incur an accuracy loss compared to the uncompressed model. In contrast, our method---also compatible with those approaches---explores the parameter space while explicitly considering the subsequent quantization, thereby better limiting performance degradation while maintaining equivalent compression capability.
Moreover, most existing DNN quantization techniques fix \textit{a priori} the quantization levels or the representative values used for weight sharing~\cite{Park17,deep_survey}. In contrast, we introduce a framework in which these representatives are learned jointly with the other DNN parameters during training. To the best of our knowledge, this is the first method that incorporates quantization parameters directly into the model as trainable variables. This is particularly relevant given recent evidence that most weights in convolutional and fully connected layers follow a bell-shaped distribution, which makes selecting appropriate quantization levels/representatives non-trivial~\cite{Li19}. 
Experimental results on the CIFAR-10 dataset using two popular architectures, AlexNet~\cite{alexnet} and VGG16~\cite{vgg16}, demonstrate the effectiveness of the proposed approach.
The paper is organized as follows: Section~\ref{sec:methods} presents the proposed methodology, including the notation used and the preliminary definitions, while Section~\ref{sec:experiments} describes the experimental results. Some concluding remarks end the paper.
\section{Methods} \label{sec:methods}
In this section, we present the notation, basic definitions, and four quantization-aware loss regularizers.
\subsection{Notation}
Matrices and vectors are written in bold, such as $\boldsymbol{W}$ and $\boldsymbol{x}$, with their entries indicated by subscripts, e.g., $w_{ij}$ or $x_i$. The symbol $:=$ is used specifically to denote definitions, distinguishing them from standard equalities.

\subsection{Preliminary Definitions}
A DNN represents a function $f_{\bW}: \mathbb{R}^{d_0} \rightarrow \mathbb{R}^{d_L}$ that maps an input vector $\boldsymbol{x} \in \mathbb{R}^{d_0}$ to an output vector $f_{\bW}(\bx) \in \mathbb{R}^{d_L}$, parameterized by $\bW$. The parameter set is $\bW = \{\bW^{(1)}, \bW^{(2)}, \ldots, \bW^{(L)}\}$, where $\bW^{(l)}$ is the weight matrix of the $l$-th layer with dimension $d_l$, for $l \in \{1, \ldots, L\}$, and $L$ is the total number of layers.
In our framework, layers may be fully connected, convolutional, recurrent, pooling, dropout, or any other type.
DNNs are inherently data-driven models designed to infer patterns from large sets of examples.
Consider a dataset $\mathcal{D} := \{(\bx_i, \by_i)\}_{i=1}^N$ of $N$ input–output pairs, where $\bx_i \in \mathbb{R}^{d_0}$ and $\by_i \in \mathbb{R}^{d_L}$.  
The objective is to learn parameters $\bW$ such that $f_{\bW}$ closely captures the mapping from inputs $\bx_i$ to their ground-truth targets $\by_i$.  
To this end, a loss function is used to evaluate performance by measuring how far the model’s predictions deviate from the true labels in the dataset.  
Formally, training consists of minimizing the loss function $\mathcal{L}$:
\begin{displaymath}
    \mathcal{L}(\bW, \mathcal{D}) := \frac{1}{|\mathcal{D}|} \sum_{(\bx, \by) \in \mathcal{D}} err(f_{\bW}(\bx), \by),
\end{displaymath}
where $err: \mathbb{R}^{d_L} \times \mathbb{R}^{d_L} \rightarrow \mathbb{R}$ denotes a task-dependent loss (e.g., cross-entropy in classification).
To compute
\begin{equation}\label{eq:loss}
    \hat{\bW} := \arg\min_{\bW} \mathcal{L}(\bW,\mathcal{D}),
\end{equation}
modern deep learning approaches typically use gradient-based optimization together with backpropagation~\cite{backprop}. Recent work has revealed that most weights in
convolutional and fully connected layers cluster around
 zero, forming an approximately bell-shaped distribution~\cite{Li19}. This distribution undermines post‑training compression via weight quantization, since it leads to large approximation errors for many weights~\cite{Park17}. To address this issue, we propose two static and two dynamic weight regularization schemes
explicitly designed to promote quantization‑friendly weight distributions.
\begin{figure*}[t]
    \centering
    \includegraphics[width=0.65\textwidth]{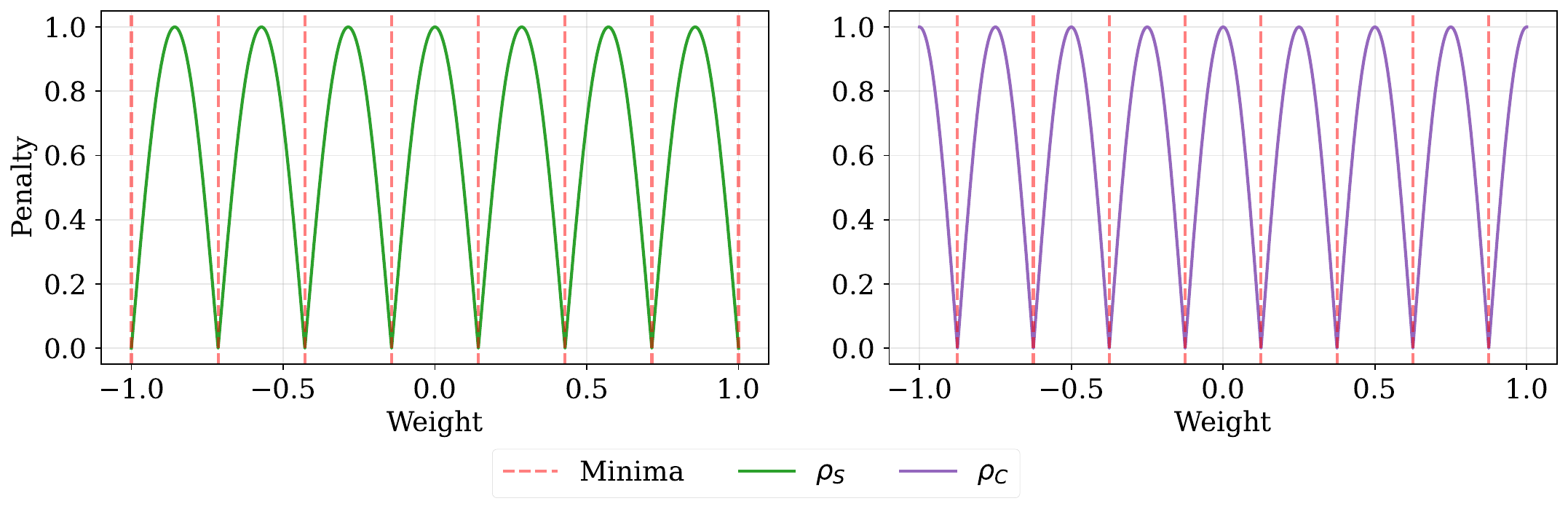}
    \caption{$\rho_S$ (left) and $\rho_C$ (right) penalty functions in the interval $[w=-1, W=1]$ for $K=8$.}
    \label{fig:reg_penalties_sinreg_cosreg}
\end{figure*}
\subsection{Quantization‑Aware Weight Regularization} \label{sec:periodic_reg}
Building on established regularization methods, we aimed to design new techniques tailored to network quantization. Our goal is a regularizer that mimics L1 penalization to support network pruning by driving some weights toward $0$ during training, enabling later magnitude-based pruning with a chosen threshold. Similarly, we hypothesized that an additional, tailored regularization term could drive weights to concentrate around preset values, rendering the network inherently ``quantization-friendly'' by the end of training. Consequently, the optimization objective extends (\ref{eq:loss}) as follows:
\begin{equation}\label{eq:loss_r}
    \hat{\bW} := \arg\min_{\bW} \mathcal{L}(\bW,\mathcal{D}) +\lambda \mathcal{R}(\bW),
\end{equation}
where $\mathcal{R}$ is 
the loss regularization term intended to create “basins of attraction’’ within a predefined weight range, whereas $\lambda > 0$ is a regularization parameter. $\mathcal{R}$ should tend to $0$ for weights located near these basins and increase as weights move away from them. This encourages weights to cluster naturally during training.
Such a mechanism can help the network explore the parameter space more effectively than unregularized training, particularly when a quantization procedure will be applied afterward.
\begin{figure*}
    \centering
    \includegraphics[width=0.65\textwidth]{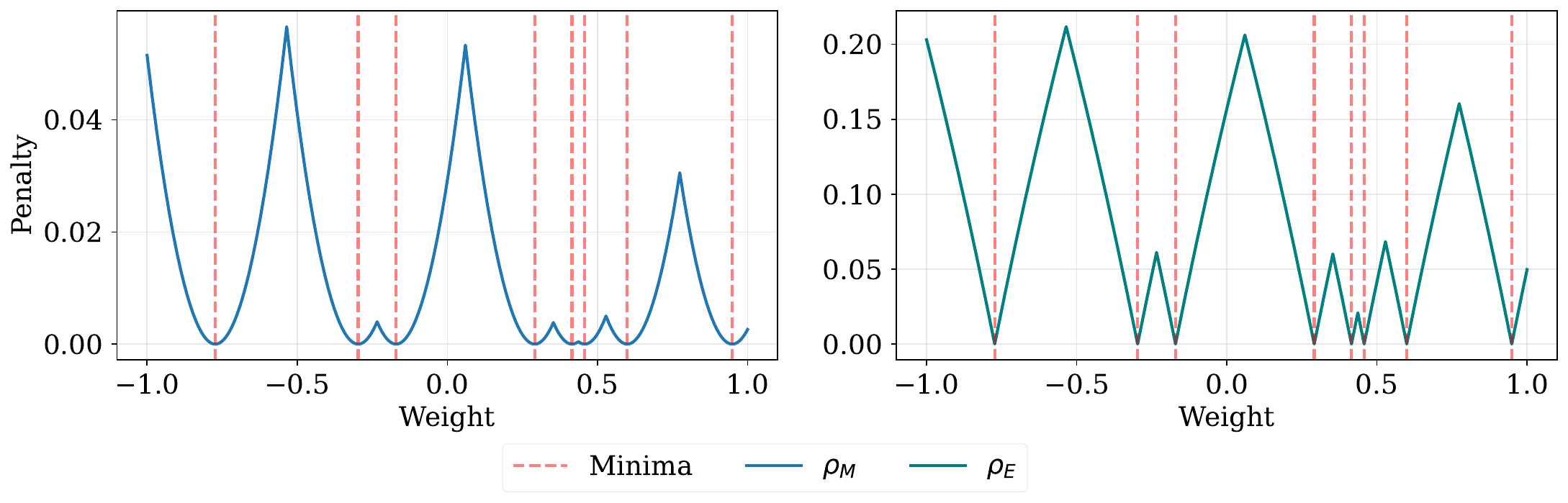}
    \caption{$\rho_M$ and $\rho_E$ functions in the interval [-1, 1] when minima ($\bu$ assignment) are those denoted by the vertical dashed lines (K=8).}
    \label{fig:reg_penalties_minl2_expreg}
\end{figure*}
\subsubsection{Periodic Quantization‑Aware Weight Regularization} \label{sec:periodic_reg}
In the first approach, we introduce two periodic regularizers based on trigonometric functions. These regularizers produce a fixed number of basins, corresponding to the minima of the regularization term, which are evenly spaced across the weight interval.
The first quantization-aware regularizer utilizes the absolute value of a sine function to create $K$ periodic minima. The sine regularizer $\mathcal{R}_{S}$ is defined as:
\begin{displaymath}
\mathcal{R}_{S}(\bW, K, w, W) :=\sum_{l = 1}^{L} \frac{\sum_{i=1}^{d_{l-1}}\sum_{j=1}^{d_l}  \rho_S(w_{ij}^{(l)}, K, w, W) 
}{d_{l-1}d_l}, 
\end{displaymath}
with
\begin{displaymath}
\rho_S(\bar w, K, w, W) := \Bigg| \sin \frac{\pi (K - 1) \cdot (\bar w - w)}{W - w} \Bigg|
\;,    
\end{displaymath}
$K$ the number of minima desired, $w$ and $W$ the minimum and maximum weight allowed, respectively. As already mentioned, since the weights in DNN typically follow a bell distribution, a priori determining $w$ and $W$ does not represent a limitation (extremes need only to not be to close). The minima are evenly distributed in $[w, W]$, and the lower $K$, the stronger the quantization we are demanding to the network, the higher the subsequent potential DNN compression (Figure~\ref{fig:reg_penalties_sinreg_cosreg}, left).
It is worth noting that, to simplify the discussion, we are assuming the weight range to be the same across layers, but clearly we can easily tailor them to the specific layer type.

Similarly, the quantization-aware regularizer $\mathcal{R}_C$ leverages the periodicity of the cosine function and is defined as follows:
\begin{displaymath}
    \mathcal{R}_{C}(\bW, K, w, W) :=\sum_{l = 1}^{L} \frac{\sum_{i=1}^{d_{l-1}}\sum_{j=1}^{d_l} \rho_C(w_{ij}^{(l)}, K, w, W)}{d_{l-1}d_l}, 
\end{displaymath}
where
\begin{displaymath}
\rho_C(\bar w, K, w, W):= \Bigg| \cos \frac{\pi K (\bar w - w)}{W - w} \Bigg|\;.
\end{displaymath}
The main difference lies in the phase of the periodic function, which shifts the locations of the static basins of attraction,
and the most external minima now do not lie exactly on $w$ and $W$ (Figure \ref{fig:reg_penalties_sinreg_cosreg}, right). 
The functions $\rho_S$ and $\rho_C$ have the advantage to be rapidly computed, since they do not require any additional parameter. However they have two main limitations: the static nature of their minima, which requires a predefined reference weight range, and the fact that these minima are evenly distributed in the interval $[w, W]$. Since the weights typically follow a bell-shaped distribution nearly centered at $0$, it would be preferable to have basins that can be adapted to the local weight distribution. To address these limitations, we propose two solutions that replace static basins of attraction with learnable ones.

\subsection{Dynamic and Learnable Quantization‑Aware Weight Regularization} \label{sec:prototype_regularization}
The rationale behind this second category of quantization‑friendly weight regularizers is to make the minima of the regularization function learnable during model training. To this end, we introduce  $K$ additional learnable parameters $\bu:=(u_1, \ldots, u_K)$, which act as adaptive basins of attraction for the weights. In this framework, instead of pulling weights toward fixed, predefined values, we encourage them to cluster around
$\bu$.  These parameters are updated at each iteration of the training algorithm, via the backpropagation procedure, just like the other model parameters $\bW$, and the regularizer penalizes weights according to their difference with $u_1, \ldots, u_K$. From this standpoint, they slightly increase the complexity with respect to $R_S$ and $R_C$ regularizers (linearly with $K$).  
The first regularizer in this category, $\mathcal{R}_M$, aims at minimizing the minimum squared difference of weights with any of the minima $\bu$. It is formally defined as it follows
\begin{displaymath}
    \mathcal{R}_{M}(\bW, \bu) := \sum_{l = 1}^{L}\frac{\sum_{i=1}^{d_{l-1}}\sum_{j=1}^{d_l} \rho_M(w_{ij}^{(l)}, \bu)}{d_{l-1}d_l}\; ,
\end{displaymath}
with
\begin{displaymath}
\rho_M(\bar w, \bu) := \min_{r \in \{1, \ldots, K\}} (\bar w - u_r)^2\;    
\end{displaymath}
By minimizing the minimum distance to minima $\bu$, we force each weight to fall in one of the clusters which will be formed during the training and relative to one representative in $\bu$. Indeed, the penalty increases quadratically with the distance from such a minimum, strongly encouraging weights to cluster tightly.
Figure \ref{fig:reg_penalties_minl2_expreg} (left) illustrates the shape of $\rho_M$ for a given assignment of $\bu$. 
The function is, in principle, unbounded, and its peaks can vary in magnitude depending on the distance from the nearest minimum. On the other side, when two minima lie very close to each other, weights falling between them may produce peaks that are too low to significantly influence the optimization process (as illustrated by the low peak in the figure).
To address this issue, the second regularization function in this category, $\mathcal{R}_E$, leverages an exponential formulation to both bound the regularization term and amplify the penalty for weights that fall between two closely spaced minima. Formally, the $\mathcal{R}_E$ weight regularization is
\begin{displaymath}
    \mathcal{R}_E(\bW,\bu) = \sum_{l = 1}^{L}\frac{\sum_{i=1}^{d_{l-1}}\sum_{j=1}^{d_l} \rho_E(w_{ij}^{(l)}, \bu)}{d_{l-1}d_l}\; ,
\end{displaymath}
where
\begin{displaymath}
\rho_E(\bar w, \bu) := 1 - \exp \left(-\min_{r \in \{1, \ldots, K\}} \; |\bar w - u_r| \right)\; .    
\end{displaymath}
The $\rho_E$ penalty is bounded between $0$ and $1$, and it approaches $1$ asymptotically as the minimum distance increases. The use of the absolute difference also ensures that weight deviations smaller than 1 have a stronger effect than in the quadratic formulation.  
As shown in Figure \ref{fig:reg_penalties_minl2_expreg} (right), for the same configuration of the $\bu$ parameters (vertical dashed lines), the peaks between closely spaced minima now exhibit a much higher relative magnitude compared with the $\rho_M$ case.
It is relevant to mention that, although the minimum function is not differentiable at every point, this does not create practical difficulties for gradient‑based optimization. The only points where the derivative is not uniquely defined are those where two or more terms achieve the same minimum value. Modern optimization frameworks handle these cases automatically by using subgradients; since such non-differentiable points are isolated, this do not affect the overall optimization dynamics.
\subsection{Quantization via Weight Sharing}
Once the weights are quantized, the storage requirements can be reduced in two main ways: either by decreasing the number of bits used to represent each weight, or by storing the representative weights in full precision and sharing them within each group through lossless structures that exploit the low entropy introduced by quantization~\cite{efficient_entropy_coding}. We focus on the weight sharing, which in our setting has the advantage of preserving the full‑precision values of the representatives, thereby better preserving the model’s accuracy. This strategy considers a representative weight in each cluster (e.g., the centroid), and then substitutes each weight in that cluster with the corresponding representative. Since our goal is to assess the effectiveness of quantization‑aware regularization, we first focus on model accuracy at a given quantization level rather than on the resulting compression rate, as lossless formats for low‑entropy weight matrices depend primarily only on the number of distinct weights used.
\section{Experiments}\label{sec:experiments}
In this section we describe the dataset and the DNNs used in the experiments, the corresponding settings and our results.

\subsection{Datasets}
We focused on the CIFAR-10~\cite{cifar10} benchmark,
containing 60K $32 \times 32$ images with three color channels (RGB), organized in 10 classes and split in 50K$+$10K images, respectively for training and testing.
Pixel values were rescaled to the range $[0,1]$ and centered by subtracting the per-channel mean.
\subsection{Deep Models}
We tested our regularization framework on two well-known and publicly available architectures: AlexNet~\cite{alexnet} (around $22$M parameters) and VGG16 \cite{vgg16} (around $33$M parameters), both adapted for the $32 \times 32 \times 3$ RGB input images. 
Except for some small adaptations to handle the different input dataset, models have been trained using the same configurations proposed in the original paper.
\subsection{Performance Assessment}
To measure the effectiveness of our quantization‑tailored regularization in facilitating subsequent weight sharing,
we trained a baseline model without regularization, applied $k$-means to the resulting weights,
and replaced weights in each cluster with the corresponding centroid~\cite{deep_survey}. We evaluated two metrics on the test set: \textit{pre‑}
and \textit{post‑tuning accuracy}, the latter measured after a few epochs of cumulative tuning, and reported as final performance assessment. In cumulative training only centroids are updated, following~\cite{Han15}.

We compared these results with those of an identical model, trained under the same conditions but using our regularization method. In this case, since weight representatives are already embedded in the regularization term, applying $k$-means is unnecessary. Clusters are instead formed by assigning each weight to the closest representative. To reduce possible distortions, 
the representative of each cluster is recomputed as the centroid of the assigned weights, which may differ slightly from the original representative specified in the regularization term (e.g., the $\bu$ values used in the $\rho_M$ and $\rho_E$ terms). The accuracy ratio measure the final performance, with values $>1$ in favor of our methods. 
\subsection{Results}
\begin{figure*}[t]
    \centering
    \includegraphics[width=0.63\textwidth]{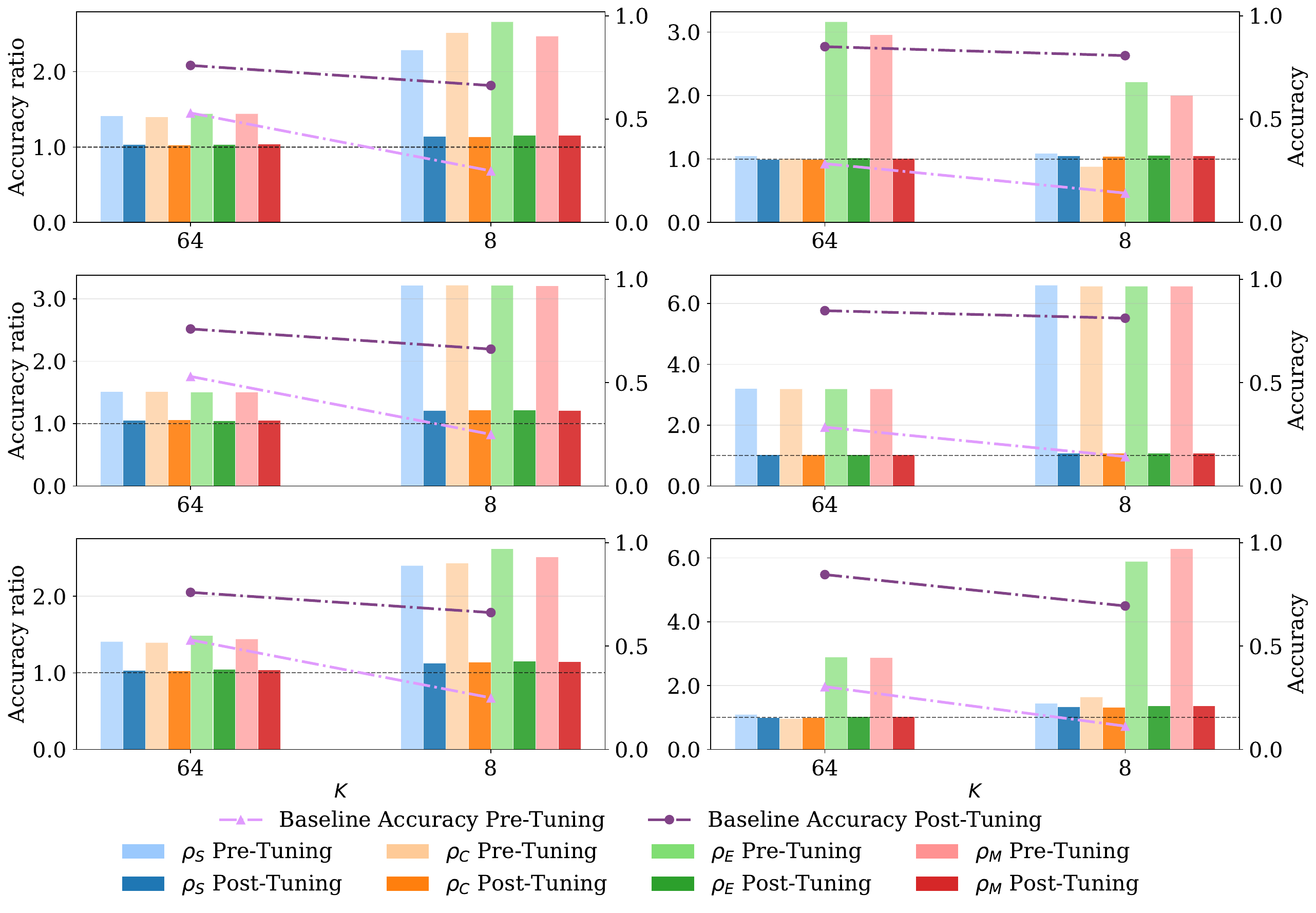}
    \captionsetup{justification=centering}
    \caption{
    Accuracy ratio for AlexNet (first column) and for VGG16 (second column) on CIFAR-10 data. Rows correspond, from top to bottom to regularizing only convolutional layers, only dense layers, and all layers. Dotted lines denote a tie: all values above mean a gain for the quantization-aware regularization. The second axis shows the baseline accuracy.}
    \label{fig:results}
\end{figure*}

To better investigate the behavior of our methodology, we conducted three experiments, regularizing (i) only convolutional layers, (ii) only the dense layers, and (iii) the entire model. We used two quantization levels ($K= 8$ and $K=64$) to assess the impact of both more and less aggressive quantization. Figure~\ref{fig:results} depicts the corresponding results. Each experiment was repeated three times with different random seeds, and the average performance computed to increase robustness.
A first clear trend that emerges is that the pre‑tuning accuracy is almost always higher when using our regularizers, up to $3\times$ on AlexNet and to $6\times$ on VGG16 better than the baseline, which, though expected, goes beyond our expectations. Conversely, the post‑tuning gain in most experiments was not something we could take for granted. In particular, for $K=8$ our regularizers always yield improvements on AlexNet, whereas on VGG16 this gain is evident only when compressing all layers (third row), likely because VGG16 is larger than AlexNet and therefore allows for stronger quantization. When $K=64$, our approach outperforms the baseline mainly in the pre‑tuning evaluation, while in the post‑tuning setting, we observe only a slight improvement on AlexNet and a performance tie on VGG16. This may depend on the fact that, in general, VGG16 can be compressed more than AlexNet.

Concerning the regularizer comparison, the dynamic family ($\mathcal{R}_M$ and $\mathcal{R}_E$) yields slightly better results on AlexNet and, in most cases, dramatically better results on VGG16 than the static counterpart, highlighting the strong potential of having learnable (and therefore adaptable) quantization levels. 
Among individual families, $\mathcal{R}_E$---when not tied with $\mathcal{R}_M$---achieves higher accuracy in $4$ out of $5$ cases, confirming our intuition that it better handles close minima. In the static family, instead, $\mathcal{R}_S$ and $\mathcal{R}_C$ achieve very similar performances.
\section*{Conclusion}
In this work, we introduced one of the first approaches that explicitly regularize neural network weights with the subsequent quantization step in mind. To the best of our knowledge, it is also the first method that learns quantization levels as model parameters, optimized jointly with the weights through backpropagation, enabling a quantization process that adapts to the loss landscape and layer characteristics.
Our experiments on AlexNet and VGG16 show substantial pre‑tuning improvements---up to $3\times$ and $6\times$ gains over the baseline, respectively. These results indicate that the proposed regularizers successfully guide the model toward weight configurations that are inherently more quantization‑friendly. Post‑tuning gains also appear, especially for aggressive quantization ($K = 8$).
Future work includes exploring weight and representative initialization strategies that exploit the typical bell‑shaped distribution of neural network weights, potentially improving the synergy between initialization, regularization, and quantization. Additionally, evaluating our approach in combination with other compression techniques, and on larger datasets and more diverse architectures, may further reveal its generality and practical impact.

\bibliographystyle{IEEETran}
\bibliography{references}

\end{document}